\documentclass[10pt,twocolumn,letterpaper]{article}

\usepackage{cvpr}              

\long\def\ignorethis#1{}

\newlength\paramargin
\newlength\figmargin
\newlength\secmargin

\usepackage{array}
\newcolumntype{L}[1]{>{\raggedright\let\newline\\\arraybackslash\hspace{0pt}}m{#1}}
\newcolumntype{C}[1]{>{\centering\let\newline\\\arraybackslash\hspace{0pt}}m{#1}}
\newcolumntype{R}[1]{>{\raggedleft\let\newline\\\arraybackslash\hspace{0pt}}m{#1}}

 \usepackage[utf8]{inputenc} 
 \usepackage[T1]{fontenc}    
 \usepackage{url}            
 \usepackage{booktabs}       
 \usepackage{amsfonts}       
 \usepackage{nicefrac}       
 \usepackage{microtype}      
 \usepackage{times}
 \usepackage{epsfig}
 \usepackage{amsmath}
 \usepackage{amssymb}
 \usepackage{adjustbox, color, paralist, enumerate, booktabs, multirow, array, tabu, float, bm}
 \usepackage{graphicx}
\usepackage{float}  
\usepackage{subfloat}
\usepackage{xcolor}
\usepackage{url} 
\usepackage{diagbox}
\usepackage{graphicx}
\usepackage[linesnumbered,ruled,vlined]{algorithm2e}
\usepackage{multirow}

\usepackage[pagebackref,breaklinks,colorlinks]{hyperref}

\usepackage[capitalize]{cleveref}
\crefname{section}{Sec.}{Secs.}
\Crefname{section}{Section}{Sections}
\Crefname{table}{Table}{Tables}
\crefname{table}{Tab.}{Tabs.}

\def\cvprPaperID{*****} 
\def\confName{CVPR}
\def\confYear{2022}

\begin{document}

\title{SparseByteNN: A Novel Mobile Inference Acceleration Framework Based on Fine-Grained Group Sparsity}


\author{Songwei Liu\thanks{Equal contribution} \quad Haitao Xu $^{*}$ \quad Yuyang Xu$^{*}$ \\ Shuai Wang \quad Jiashi Li  \quad Chenqian Yan  \quad Liangqiang Li \quad Lean Fu \quad Xin Pan  \quad Fangmin Chen \\
ByteDance Inc \\
{\tt\small \{liusongwei.zju, xuhaitao.henry, xuyuyang.nicole, wangshuai.hust, lijiashi\}@bytedance.com}\\ 
{\tt\small \{yanchenqian.i, lilianqiang, fulean, panxin.321\}@bytedance.com}\\ 
{\tt\small cfangmin@gmail.com}
}
\maketitle

\begin{abstract}
\vspace{-0.25em} 
To address the challenge of increasing network size, researchers have developed sparse models through network pruning. However, maintaining model accuracy while achieving significant speedups on general computing devices remains an open problem.  In this paper, we present  a novel mobile inference acceleration framework SparseByteNN, which leverages fine-grained kernel sparsity to achieve real-time execution as well as high accuracy. Our  framework consists of two parts: (a) A fine-grained kernel sparsity schema with a sparsity granularity between structured pruning and unstructured pruning. It designs multiple sparse patterns for different operators. Combined with our proposed whole network rearrangement strategy, the schema achieves a high compression rate and high precision at the same time. (b) Inference engine co-optimized with the sparse pattern. The conventional wisdom is that this reduction in theoretical FLOPs does not translate into real-world efficiency gains. We aim to correct this misconception by introducing a family of efficient sparse kernels for ARM and WebAssembly. Equipped with our efficient implementation of sparse primitives, we show that sparse versions of MobileNet-v1 outperform strong dense baselines on the efficiency-accuracy curve.  Experimental results on Qualcomm 855 show that for 30\% sparse  MobileNet-v1, SparseByteNN achieves 1.27$\times$ speedup over the dense version and 1.29$\times$ speedup over the state-of-the-art sparse inference engine MNN with a slight accuracy drop of 0.224\%. The source code of SparseByteNN will be available at https://github.com/lswzjuer/SparseByteNN.

\end{abstract}

\section{Introduction}
\label{sec:intro}
 \begin{figure}[ht]
	\centering
	\includegraphics[width=0.9\linewidth]{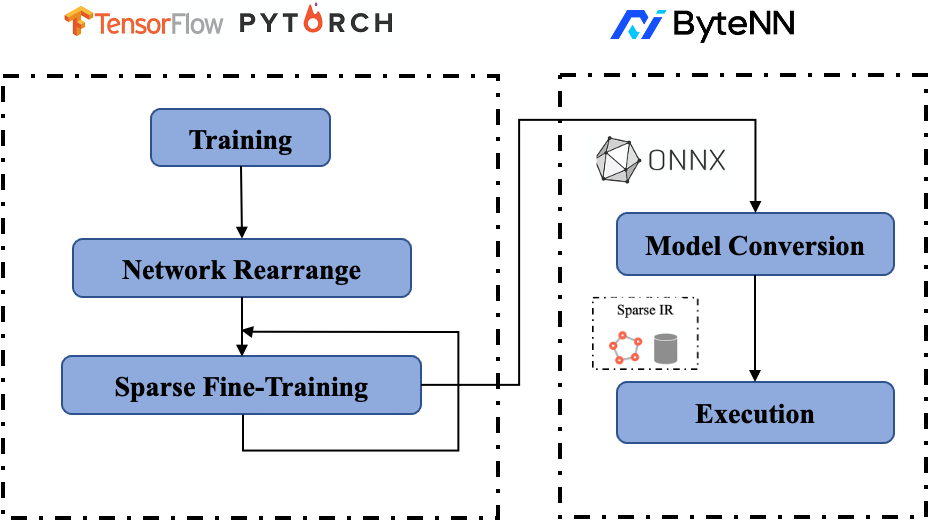}
	\caption{SparseByteNN overview}
	\label{framwork}
\end{figure}

Deep convolutional neural networks (CNNs) have achieved extraordinary performance in  computer vision tasks and  become the fundamental element and core enabler of ubiquitous artificial intelligence.
With the fast growth of embedded and mobile applications,  executing CNNs on mobile platforms is becoming increasingly attractive, which will improve computing power utilization, enhance data security, and reduce dependence on the network\cite{deng2019deep}\cite{lane2016deepx}\cite{lane2017squeezing}. However, typical state-of-the-art(SOTA) CNNs models are computation-extensive and memory-hungry. Even mobile devices with advanced CPUs and GPUs are considered resource-constrained when executing them. Thus, achieving efficient inference with real-time performance is still a challenging task.

To achieve this goal, extensive efforts have been made for the optimization of algorithms, software, and hardware. Algorithm optimization includes efficient network backbone design and  model compression. Early SOTA  CNNs\cite{vgg}\cite{resnet}\cite{inceptionv4}usually have a backbone stacked by normal 3$\times$3 convolution (CONV) layers. Their computationally prohibitive cost makes real-time deployment on mobile devices almost impossible.  \cite{mbv1}\cite{mbv2}\cite{ghostnet}\cite{efficientnet} use depthwise separable convolution instead of normal CONV layers to build efficient models for mobile and embedded vision applications, which become the mainstream of mobile network design. In order to further reduce the redundancy of CNNs,  model compression techniques, including model pruning\cite{deep}\cite{learning}\cite{agp}\cite{l1}\cite{l2}\cite{slim}\cite{fpgm}\cite{tylor} and model quantization\cite{gupta2015deep}\cite{courbariaux2016binarized} have been proposed and studied intensively for model storage reduction and computation acceleration. Weight quantization is less supported in mobile devices, especially mobile GPUs\cite{patDNN}. Therefore, this paper leverages model pruning as the primary model compression technique. Recent developments in pruning can be mainly divided into weight pruning\cite{deep}\cite{learning}\cite{agp} and filter pruning\cite{l1}\cite{tylor}\cite{slim}\cite{fpgm}. Weight pruning directly removes weight values at any position in the network, which is demonstrated to achieve an extremely high compression rate with high accuracy performance. However, weight pruning is not friendly for hardware or software optimization. Specifically, the compression makes few contributions to memory access saving and calculation acceleration on general CPU(SIMD) and GPU(SIMT) architecture. In contrast, filter pruning directly 
removes the entire filter in the convolutional neural network, which can generate hardware-efficient regular models but fails to maintain accuracy beyond moderate sparsity ratios.  Especially for mobile-oriented lightweight CNNs, such as Mobilenet\cite{mbv1}, due to the small redundancy of model parameters, filter pruning encounters severe accuracy loss problems.

We notice that the pruning granularity of weight pruning and filter pruning represent two extremes in the design space, leading to the failure of balancing model accuracy and speedup gains. Besides, these optimization algorithms are isolated and have not been co-optimized with software and hardware optimization. In this paper, we introduce a new pruning strategy called fine-grained kernel group pruning(FKGP), whose sparsity granularity is between weight pruning and structured pruning, revealing a previously unknown point in the design space.  In particular, for the core operators in the mobile network, including pointwise convolution(Conv1$\times$1) and depthwise convolution(DwConv3$\times$3), we designed diverse sparse patterns, which can have a better trade-off between accuracy and hardware efficiency.
Our fine-grained kernel sparsification is implemented in groups, which means that kernels in the same group are kept or removed uniformly, and kernels in the kept group have the same  sparse pattern. Compared with single kernel sparse, group kernel sparse has less precision loss but is more friendly to parallel acceleration. Based on this, we propose a whole network rearrangement strategy to derive a more influential kernel group for accuracy improvements.  The above fine-grained sparse patterns cannot be directly accelerated by a general inference engine, so we introduce a family of efficient sparse kernels for ARM and WebAssembly to translate reduction in theoretical FLOPs to hardware efficiency. 

In summary, we propose a novel end-to-end mobile acceleration framework named \emph{SparseByteNN}. Combined with the improved algorithm optimization strategy and sparse engine implementation,  \emph{SparseByteNN} advances SOTA in model pruning and open source Inference engine. The overall framework of \emph{SparseByteNN} is shown in Fig \ref{framwork}. Our contributions can be summarized as follows:


1.  We focus on the acceleration of mobile lightweight CNNs,  and design fine-grained kernel group sparse strategies for Conv1$\times$1 and DwConv3$\times$3 respectively. The co-optimized sparse patterns achieve an extremely high compression rate with high accuracy performance. Moreover, with the high-performance sparse kernel implementation for ARM and WebAssembly, the designed patterns can recover the hardware efficiency lost due to the fine-grained patterns. For Conv1$\times$1, we demonstrate a geometric mean of speedups of 26.80\% compared to the dense network at 30\% sparsity. In particular, we achieve high-performance compression of DwConv3$\times$3,  which can speed up by up to 49.6\% at 33\% sparsity.

2.  We propose a  whole network rearrangement strategy, which divides kernels with similar importance into a group, improves the accuracy of each group's importance evaluation and derives a more influential kernel group for accuracy improvements.

3.  We propose an end-to-end model acceleration framework  \emph{SparseByteNN}, consisting of three components: a) \emph{compression algorithm component}, which provides out-of-the-box pruning capabilities for pre-trained models  b)  \emph{model conversion tool}, which converts the model IR of the training framework into Model IR of sparse engine c)\emph{sparse inference engine}, which provides efficient inference implementation compatible with CPUs for fine-grained kernel group sparsity.

\section{Related Works}
\begin{figure}[t]
	\centering
	\begin{subfigure}{1\linewidth}
		\centering
		\includegraphics[width=1\linewidth]{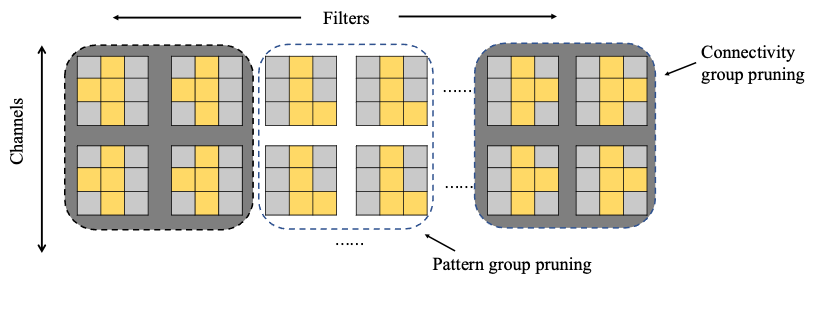}
		\caption{Connectivity group pruning and 5:9 pattern group pruning for Conv3x3}
		\label{59}
	\end{subfigure}
	
	\centering
	\begin{subfigure}{1\linewidth}
		\centering
		\includegraphics[width=1\linewidth]{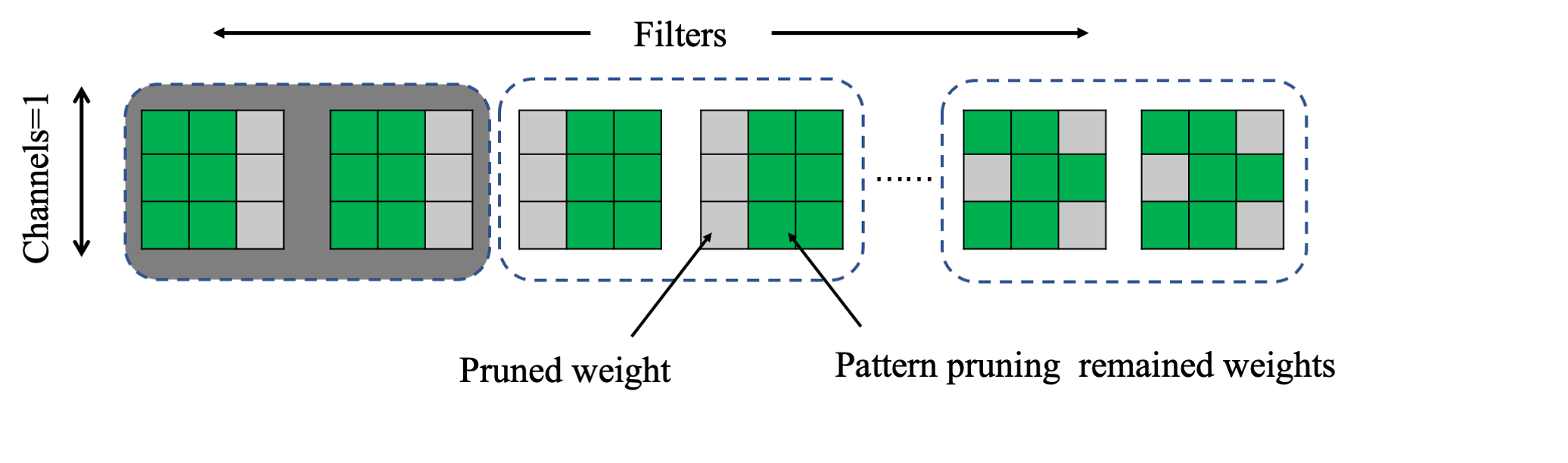}
		\caption{ 3:9 pattern  group pruning for DwConv3x3}
		\label{39}
	\end{subfigure}
	
	\centering
	\begin{subfigure}{1\linewidth}
		\centering
		\includegraphics[width=1\linewidth]{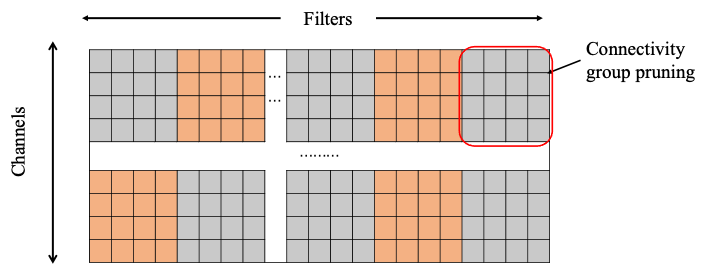}
		\caption{Connectivity group pruning  for Conv1x1}
		\label{44}
	\end{subfigure}
	\caption{Illustration of the implementation form of fine-grained kernel group sparsity on core operators}
	\label{all_patterns}
\end{figure}
\subsection{Model Pruning}
The improvement of neural network performance is usually accompanied by the increase of resource requirements  such as params and flops,  One popular approach for reducing them at test time is model pruning, which can be categorized into weight pruning and filter pruning.  Weight pruning dates back to Optimal Brain Damage\cite{lecun1989optimal}, which prunes weights based on the Hessian of the loss function. Many recent works\cite{deep}\cite{learning}\cite{agp} have further optimized the pruning evaluation criteria and pruning methods. For example,  Han et al.\cite{learning}  proposed a three-step strategy including training, pruning, and fine-traing to remove unimportant connections and restore accuracy. Michael et.al\cite{agp} proposed a gradual pruning technique that can be seamlessly incorporated into the training process. Although it is an adaptive in-training pruning strategy, it cannot  recover from premature pruning. Lin et al.\cite{lin2020dynamic} proposed a dynamic allocation of sparsity patterns and incorporated feedback signals to reactivate prematurely pruned weights.
Weight pruning  focuses on pruning the fine-grained weight of filters leading to unstructured sparsity in models, which cannot be directly accelerated on general computing libraries.  In contrast, filter pruning targets pruning the entire filter, which could achieve structured sparsity. \cite{apoz} proposed to explore sparsity in activations
for network pruning.  \cite{l2} uses l2-norm to select unimportant filters and explores the sensitivity of layers for filter pruning. \cite{slim}introduces sparsity on the scaling parameters of batch normalization (BN) layers to prune the network. \cite{tylor}  proposes a Taylor expansion-based pruning criterion to approximate the change in the cost function induced by pruning. To reduce dependence on pre-trained models and improve model capacity, \cite{he2018soft}\cite{fpgm}proposed soft filter pruning  enables the pruned filters to be updated when training the model after pruning.  Although the pruned model obtained by filter pruning can take full advantage of high-efficiency Basic Linear Algebra Subprograms (BLAS) libraries to achieve better acceleration but fails to maintain accuracy beyond moderate sparsity ratios. The pruning granularity of weight pruning and filter pruning represent two extremes in the design space, causing them to fail to balance accuracy and acceleration gains. 

Recently, some  work has noticed this problem and proposed some pattern-based or block-based weight pruning schemes  with compiler-based optimizations \cite{patDNN}\cite{admm}\cite{24}. Similar to our work, their pruning granularity is between weight pruning and filter pruning  to balance  accuracy and inference. \cite{24} describe a  2:4 pattern pruning  scheme and  NVIDIA Ampere architecture  introduces Sparse Tensor Cores to provide dedicated acceleration capabilities for this sparse mode. Furthermore, PatDNN\cite{patDNN}  uses Alternating Direction Methods of Multipliers(ADMM) and  pattern-based weight pruning schema  to solve a fine-grained  sparse model and performs compiler optimizations to achieve real-time mobile inference. PatDNN mainly optimized the performance of  Conv3x3, but the principal layers of the mobile network represented by MobileNet-v1\cite{mbv1}  are Conv1$\times$1 and DwConv3$\times$3, which means that it suffers difficulties when generalized to mobile networks.  In contrast, SparseByteNN focuses on the optimization of mobile networks, and designs customized 4$\times$4 and 16$\times$1 pattern-based sparsity for Conv1$\times$1  and DwConv3$\times$3 respectively, and replaces compilation optimization with expert-level manual optimization, achieving a more extreme performance.

\subsection{Acceleration Frameworks on Mobile}

On-mobile neural network deployment relies on the performance of inference framework, so on-mobile DNN inference frameworks have attracted more and more attentions\cite{luo2020comparison}. Representative DNN acceleration frameworks, such as TensorFlow-Lite\cite{tflite}, Pytorch-Mobile\cite{pytorch-mobile}, and TVM\cite{TVM} are designed to support inference acceleration of dense neural networks. Although these inference frameworks already incorporate several graph optimization and compilation optimization strategies, including layer fusion, constant folding and Auto-Tuning, they lack the ability to further accelerate sparse models. Similar to our work, MNN\cite{Mnn} recognizes the potential of sparse speedup and supports block-based sparse speedup based on expert hand-crafted optimization, with a sparse granularity of N$\times$1. In order to improve the optimization efficiency, PatDNN\cite{patDNN} and Auto-PatdNN\cite{auto-patdnn} realize the sparse model acceleration based on compiler-based optimization. Although these frameworks support sparse acceleration, they support limited types of sparse operators and suffer difficulties when generalized to DNN layers other than Conv3×3 layers(PatDNN) and Conv1x1 layers(MNN). In Section \ref{4.2}, we will discuss this issue and compare performance.

\section{Method}
In this section, we first introduce the mathematical representation of FKGP in Section \ref{3.1}. Then we introduce the sparsity patterns of Conv3$\times$3, Conv1$\times$1, and DwConv3$\times$3 in Section \ref{3.2}, and the co-optimized implementation in Section \ref{3.3}. In Section \ref{3.4}, we describe a whole network rearrangement strategy, which can improve the performance of the sparse model. Finally, we introduce the overall framework of SparseByteNN in Section \ref{3.5}.

\subsection{Preliminaries}
\label{3.1}
For an L-layer pre-trained  model, the weights and biases for the i-th layer are denoted by $W^{i} \in\mathbb{R}^{ n^{i} \times kh^{i}  \times kw^{i} \times c^{i}} $ and $B^{i}\in\mathbb{R}^{ n^{i}}$, where $n^{i}$, $kh^{i}$, $kw^{i}$ and $c^{i}$ stand for the output channel, input channel, kernel height, and kernel width respectively. The input for i-th layer is denoted by $INPUT^{i} \in\mathbb{R}^{ ih^{i} \times iw^{i} \times c^{i} }$, where $ih^{i}$, $iw^{i}$ stand for the input height and input width. To obtain a sparse model, a general approach is to prune part of $W^{i}$, i.e., to set them to zero. This process can be implemented by applying a mask $M^{i}\in\{0,1\}$ to the weights, resulting in a sparse model $\tilde{W}^{i} = W^{i} \odot M^{i}$.  The quality of  pruning is defined as the parameter $\delta\in[0,1]$ such that  $\delta = \frac{\Vert W^{i}- \tilde{W}^{i} \Vert}{ W^{i}}$. Pruning without information loss corresponds to  $W^{i} =  \tilde{W}^{i}$ , i.e., $\delta = 0$. Thus, the pruning problem can be summarized as minimizing the $\delta$ at the pruning ratio $\rho$ with the optimal mask,
\begin{eqnarray}	
	\underset {M^{i}} { \operatorname {arg\,max} } \, \Vert W^{i} \odot M^{i} \Vert, \quad s.t. \frac{\Vert M^{i} \Vert_{0}}{K}= 1 - \rho
\end{eqnarray}
For weight pruning, the weights could be removed at random locations. In this case, the mask tensor $M^{i}$ has the same shape as $ W^{i}$ of  $\mathbb{R}^{ n^{i} \times kh^{i}  \times kw^{i} \times c^{i}} $, which is $K=n^{i} \times kh^{i}  \times kw^{i} \times c^{i} $. For filter pruning, the sparse granularity is the entire  filter. Thus, each mask $M^{i}$  has the shape of $\mathbb{R}^{ n^{i} } $ and  $K=n^{i} $.  
To facilitate the implementation of pattern-based pruning, we reformat the expression of   $W^{i} \in\mathbb{R}^{ n^{i} \times kh^{i}  \times kw^{i} \times c^{i} } $  to  $\overline{W}^{i} \in\mathbb{R}^{ n^{i} \times c^{i} } $.  Each member  in $\overline{W}^{i}$ represents a kernel  of shape $n^{i} \times c^{i}$. Semantically, the kernel is a connection channel between the input feature map and the output feature map.  For our FKGP,  we further group the kernels on the input channel $ c^{i}$ and output channel $ n^{i}$ as a whole, which are simultaneously sparse or removed.  Continuing the definition of PatDNN\cite{patDNN}, we define the two cases of fixed pattern sparse and complete removal as pattern group pruning and connectivity group pruning,  such that 
\begin{small}
\begin{equation}
	\begin{aligned}
	\underset {M^{i}} { \operatorname {arg\,max} } \,  \sum\limits_{i=0}^{\frac{n^{i}}{go}}\sum\limits_{j=0}^{\frac{c^{i}}{gi}} &\Vert \overline{W}^{i}_{i*go:(i+1)*go, j*gi:(j+1)*gi} \\
	& \odot M^{i}_{i*go:(i+1)*go, j*gi:(j+1)*gi,:,:} \Vert  \\ 
	& s.t. \frac{\Vert M^{i} \Vert_{0}}{K}= 1 - \rho
\end{aligned}
\end{equation}
\end{small}
where $go$ and $gi$ represent output channel and input channel group size respectively,  and $\rho$  is the sparsity rate.

\subsection{Fine-grained Kernel Group Sparsity}
\label{3.2}

As shown in Fig  \ref{all_patterns}, our proposed FKGP strategy designs a customized sparse strategy for Conv3x3, Conv1x1, and DwConv3x3.

\textbf{Conv3x3} is less computationally efficient than depthwise separable convolution, so it is not the core operator of mobile lightweight CNNs.  For example,  MobileNet-v1\cite{mbv1}  contains  one layer of Conv3x3, and its calculation amount is only 1.91\%.  We focus on the acceleration of  mobile lightweight CNNs so that the sparse mode of Conv3x3 is not carefully designed but directly adopts the 5:9 sparse mode proposed by PatDNN\cite{patDNN}.  As shown in Fig \ref{59}, each kernel is either completely removed called connectivity pruning, or partially removed, and the remaining weights
form specific kernel patterns called pattern pruning. Every kernel reserves 4 non-zero weights out of the original 3 $\times$ 3 kernel, which contains the central weight. PatDNN\cite{patDNN} elaborates on kernel patterns with more details.
In order to better balance speed and accuracy, we regard $g_{i} \times g_{o}(4\times4)$  kernels as a group, and each group is considered as a whole.

 \begin{table}[!htbp]
 \centering
  \caption{Sparse model accuracy under different group size configurations and the accuracy of the baseline model is 72.634\%.}
 \label{tab::conv1x1_group_num}  
 \resizebox{\linewidth}{!}{
 \begin{tabular}{c c c c c }
 	\toprule  
 	\diagbox{Acc}{ $g_{i} \times g_{o}$} & 2x2 & 4x4 & 8x8 & 16x16 \\
 	\midrule  
 	Top1-Acc(\%) &72.658 & 72.488 &72.090 &72\\
 	\midrule  
 	Top1-Acc $\uparrow$ (\%) &+0.024 & -0.146 & -0.544 &-0.634\\
 	\bottomrule 
 \end{tabular}
 }
 \vspace{-0.5em}
\end{table}

 \begin{table}[!htbp]
 \centering
  \caption{Sparse model accuracy under different group size configurations and the accuracy of the baseline model is 72.634\%.}
 \label{tab::dw_group_num}  
 \resizebox{\linewidth}{!}{
 \begin{tabular}{c c c c c }
 	\toprule  
 	\diagbox{Acc}{ $g_{i} \times g_{o}$} & 1x4 & 1x8 & 1x16 & 1x32 \\
 	\midrule  
 	Top1-Acc(\%) &72.838 & 72.866 &72.954 &72.896\\
 	\midrule  
 	Top1-Acc $\uparrow$ (\%) &+0.204 & +0.232 & +0.320 &+0.262\\
 	\bottomrule 
 \end{tabular}
 }
 \vspace{-0.5em}
\end{table}

\textbf{Conv1x1} and Fc layer are commonly transformed into GEMM, i.e., the multiplication of a weight matrix and an input matrix. Each kernel of these layers contains only one weight, and only connectivity sparsity exists in these layers.
As shown in Fig \ref{44},  we divide the weight tensor into $\frac{n^{i}}{g_{o}}\times \frac{c^{i}}{g_{i}}$ blocks with equal size($g_{i} \times g_{o}$) and apply connectivity group pruning.  The importance of each block is evaluated by $l_{1}$-norm and the $\frac{n^{i}}{g_{o}}\times \frac{c^{i}}{g_{i}}\times \rho$ blocks with the lowest importance are removed. The value of $g_{i} \times g_{o}$ needs to comprehensively consider the model accuracy and acceleration friendliness. The larger the value is, the less sparse patterns exist in the model, which is not conducive to maintaining accuracy but more conducive to acceleration. We perform 30\% connectivity group pruning on Conv1x1 in MobileNet-v1(ImageNet) to obtain accuracy with different group sizes. As shown in Table \ref{tab::conv1x1_group_num}, there is only a slight loss of accuracy when the group size is no larger than $4\times4$. When the group size is further increased to $8\times8$, the accuracy loss increases by 3.72 times. Since the group size of $4\times4$ is also computationally friendly (discussed in Section \ref{3.3}), we finally set the group size of Conv1x1 to $4\times4$.

\textbf{DwConv3x3} is one of the components of depthwise separable convolution, which is difficult to compress. Due to the loss of accuracy, the previous similar work\cite{patDNN} did not realize the pattern-based pruning of the DW layer. On the contrary, we propose 3:9  sparse patterns for the DwConv3x3 layer, which can achieve near-lossless pattern pruning. As shown in Fig \ref{39},  each kernel removes 3 weights from the original $3 \times 3$ kernel, which are taken from the first and third columns and distributed in three rows, in which case there are $2^{3}$ potential kernel patterns.  We regard $g_{i} \times g_{o}$  kernels as a group and each group selects the best kernel pattern by maximizing the $l_{1}$-norm after sparse. It should be noted that the input channel of depthwise convolution is equal to 1, so a single kernel is essentially the entire filter, which means that the connectivity pruning will degenerate into filter pruning. In order to maintain accuracy, we only perform pattern group pruning for DwConv3x3, resulting in 33\% sparsity. As
For DwConv3x3, the calculation principle determines that $g_{i}$ is fixed at 1. To determine the best $g_{o}$ value, we study the impact of different $g_{o}$ on the accuracy when only pruning the DwConv3x3. As shown in Table \ref{tab::dw_group_num}, 
DwConv3x3 pruning is not sensitive to the group size. Considering the calculation friendliness, we set  $g_{o}$ equal to 16. 

\begin{figure}[t]
	\centering
	\includegraphics[width=0.9\columnwidth]{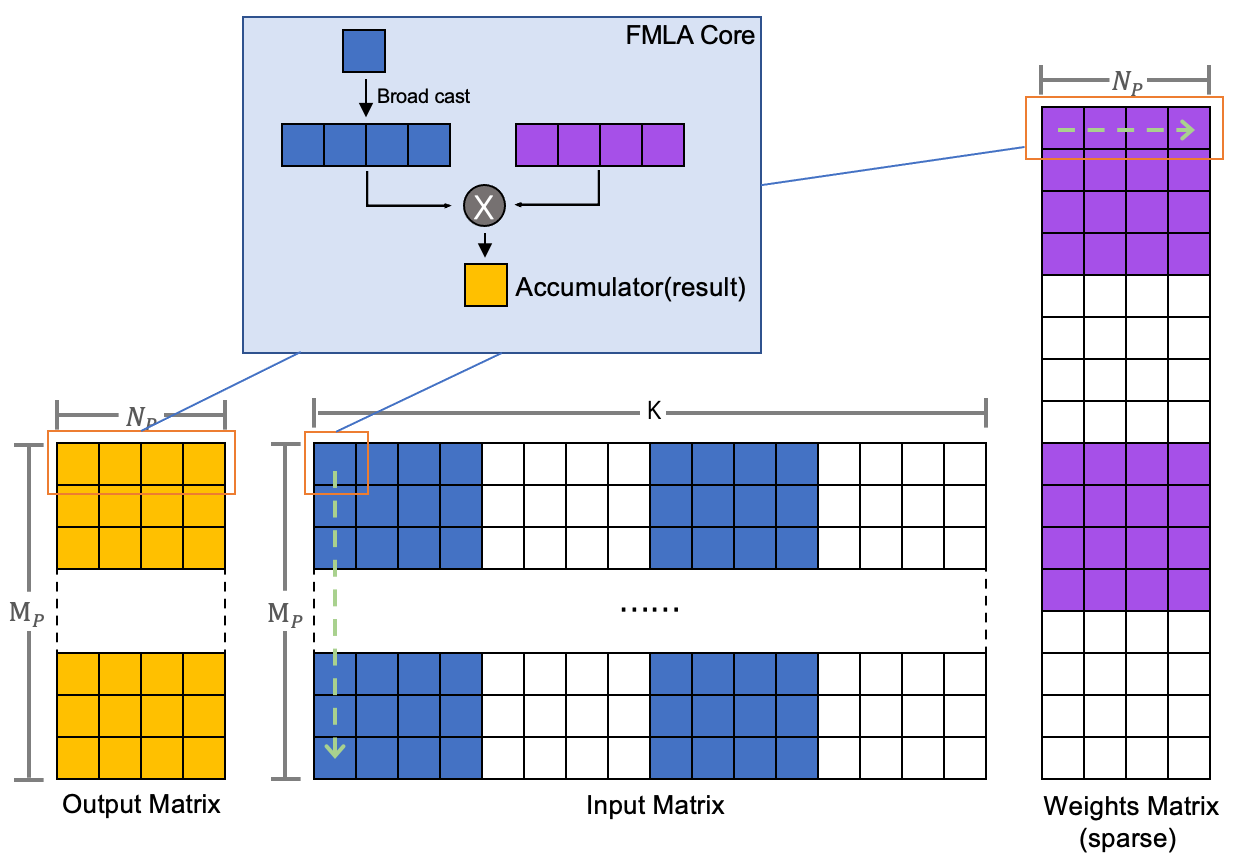}
	\caption{ Calculation flow of block-based pruning for Conv1x1}
	\label{47}
\end{figure}

\subsection{Co-design Inference Engine}
\label{3.3}
 Unstructured Conv1x1 pruning provides unique advantages in accuracy compared with structured sparsity. However, the discontinuous weights pose a problem in vectorized parallel computing and lead to increased cache misses. The random connectivity in both $n$ and $c$ dimensions results in negligible or even negative performance effects due to irregular memory accesses. To guarantee the effectiveness of random pruning, this paper propohttps://www.overleaf.com/project/63b6d9cf9a3aed58c82f50fcses a half-structured method with block sparsity units, ensuring a certain degree of continuity and effectiveness in both n and m dimensions. On one hand, the half-structured method substantially avoids the reduction in performance caused by completely random weights. On the other hand, the random with the block-sparsity level can reduce the loss of training accuracy compared to structured pruning.
 
\begin{figure}[t]
	\centering
	\begin{subfigure}{0.9\linewidth}
		\centering
		\includegraphics[width=0.9\linewidth]{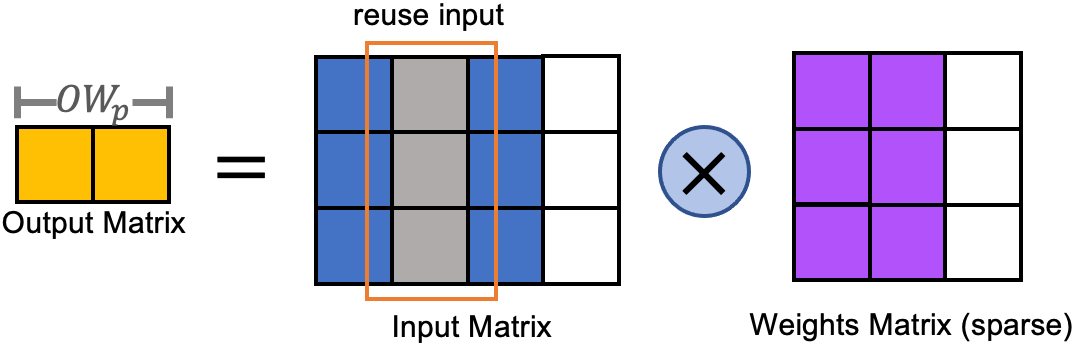}
		\caption{Weight sparsity for the third column reuses input data twice for every two output pixels}
		\label{69}
	\end{subfigure}
	
	\centering
	\begin{subfigure}{0.9\linewidth}
		\centering
		\includegraphics[width=0.9\linewidth]{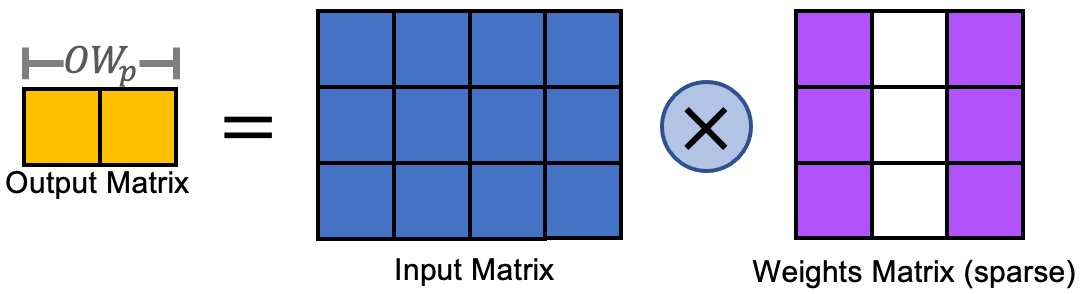}
		\caption{Weight sparsity for the middle column needs more memory access to input data}
		\label{70}
	\end{subfigure}
        \caption{Calculation method of pattern-based pruning for DwConv3x3}
	\label{depthwise_sparsity}
\end{figure}

The ARM instruction sets on mobile devices have a fixed 128-bit vector length. To adapt to hardware limitations and reduce network training loss, this study selects 4$\times$4 pixels block as minimized sparsity unit for Conv1x1 as shown in Fig \ref{44}. The performance bottleneck of mobile devices generally depends on memory access, especially on low-end devices. Therefore, the design of computing block size for Conv1x1 needs to minimize the memory access times with hardware limitations. The tiling size for input wh-dimension and output n-dimension needs to satisfy the inequality:
\begin{small}
\begin{equation}\label{ineql}
\begin{aligned}
    \min{\frac{M}{M_{p}}}\frac{N}{N_{p}}(KM_{p}+KN_{p}+M_{p}N_{p}) \\ s.t. KM_{p}+KN_{p}+M_{p}N_{p} <= R
\end{aligned}
\end{equation}
\end{small}
where $M=ih*iw$, $M_{p}$ is the block size of wh-dimension, $N_{p}$ is n-dimension block size, $K$ is the input channel, $R$ is the number of registers for ARM architecture which is 32 for armv8 and 16 for armeabiv7 in general.

\begin{algorithm}[ht]
\caption{Block-size Sparsity of Conv1x1}\label{Algorithm:Conv1x1}
\KwData{Weight $W\in\mathbb{R}^{ oc \times kh  \times kw \times ic
}$, Input feature map $I \in\mathbb{R}^{ n \times ih  \times iw \times ic}$ and sparse info SD }
\KwResult{Output feature $O \in\mathbb{R}^{ n \times oh  \times ow \times oc}$}
\emph{Set block size corresponding to the $ih*iw$ dimension and $oc$ dimension to $M_{p}$ and $N_{p}$}\;
\For{$i\leftarrow 1$ \KwTo $ih*iw/M_{p}$}{
    \For{$j\leftarrow 1$ \KwTo $oc/N_{p}$}{
    \emph{computing output block $O_{i,j}$ of $M_{p} \times N_{p}$}\;
        \For{$kIndex\leftarrow 1$ \KwTo $\in ic/4$}{
            \emph{$kStartIndex$ = SD[$kIndex$]}\;
            \emph{computing every 4 input channels as a summation factor for final result for output block data $M_{p} \times N_{p}$}\;
        }
    }
}
\end{algorithm}

The $M_{p}$ and $N_{p}$ are obtained as 20 and 4 respectively by solving the inequality. For each cycle, the calculation for every $20\times4$ output results require the input size of $20 \times K$, and the weight size of $K \times 4$. In this case, the output, input, and weight occupy 20, 5, and 4 registers respectively. 29 registers are utilized which is close to the maximum number of registers supported by the hardware. The computation flow is shown as Algorithm \ref{Algorithm:Conv1x1}.

Depthwise is another essential operator in lightweight networks and contains critical information. Inspired by PatDNN\cite{patDNN}, this paper proposes a flexible sparsity method for DwConv3x3 pruning to reduce loss of training accuracy. In our study, the first and third pixels of each row of the kernel were randomly pruned. From the perspective of memory access, the input data corresponding to the middle position of weights can be reused twice for every two output data as shown in Fig \ref{69}. However, the pruning of the middle pixel as shown in Fig \ref{70} could not reuse the input data which increases the time consumption for memory access to cache and DDR. Every 8 sparsity patterns in our study reduce the access memory by 25\% compared to the sparsity method for pruning the middle pixel.

The pseudo-code of Depthwise computing process is shown in Algorithm \ref{Algorithm:Depthwise}. The calculation in this study uses a sliding window method, and the input data format will be packed as NHWC16 to adapt to the computation for sparsity pattern. We regard 16 as a group for the output channel. The output channels that are not multiples of 16 are set to 4 or 8 as a group. In addition, to minimize the training loss of Depthwise, this study further adds a full-1 pattern, namely dense mode. The number of output channels for full-1 pattern can be dynamically increased according to the specific network training accuracy and which still follows the 16-block rules. 

\begin{algorithm}[ht]
\caption{Block-size Sparsity of Depthwise}\label{Algorithm:Depthwise}
\KwData{Weight $W\in\mathbb{R}^{ oc \times kh  \times kw \times ic
}$, Input feature map $I \in\mathbb{R}^{ n \times ih  \times iw \times ic}$ and PatternInfo}
\KwResult{Output feature $O \in\mathbb{R}^{ n \times oh  \times ow \times oc}$}
\emph{Set block size corresponding to the $ow$ dimension $ow_{p}$, and $N=oc$ $N_{p}=16,8,4$}\;
\For{$nBlock\leftarrow 1$ \KwTo $N/N_{p}$}{
    \emph{$curSparsityPattern$ = (PatternInfo)[$nBlock$]}\;
    \If{$curSparsityPattern == 0$}{
        \For{$ohIndex\leftarrow 1$ \KwTo $oh$}{
            \For{$owBlock\leftarrow 1$ \KwTo $\in ow/ow_{p}$}{
                \emph{computing every $N_{p}$ channels for output block $OW_{p} \times N_{p}$}\;
            }
        }
    }\;
    \If{$curSparsityPattern == 1$}{
        \emph{$.....$}\;
    }\;
    \emph{$.....$}\;
    \If{$curSparsityPattern == 7$}{
        \emph{$.....$}\;
    }\;
}
\end{algorithm}


In this study, the output data of 2x16 block is calculated in every computing cycle. The number of 2 and 16 stands for feature map dimension and output channel respectively. Considering the usage of registers, as shown in Fig \ref{69} the weight data requires 6x16 block size which occupies 12 neon registers. The input data requires 9x16 block size which occupies 18 neon registers. The output data is 2x16 which occupies 4 neon registers. We reuse two registers for the input data and keep the data individual from the others.

For the pruning of Conv3x3 operator, this study applies the method proposed in paper PatDNN\cite{patDNN} that 56 sparsity patterns were implemented. The same sparsity pattern will be shared by several adjacent filters which can be dynamically selected by the number of 4, 2, and 1 during network training considering the balance between training accuracy and performance. In this paper, we mainly focus on  weights pruning for lightweight networks, so the sparsity of Conv3x3 operator will not be introduced in detail.

\begin{figure}[t]  
	\centering
	\includegraphics[width=0.7\textwidth]{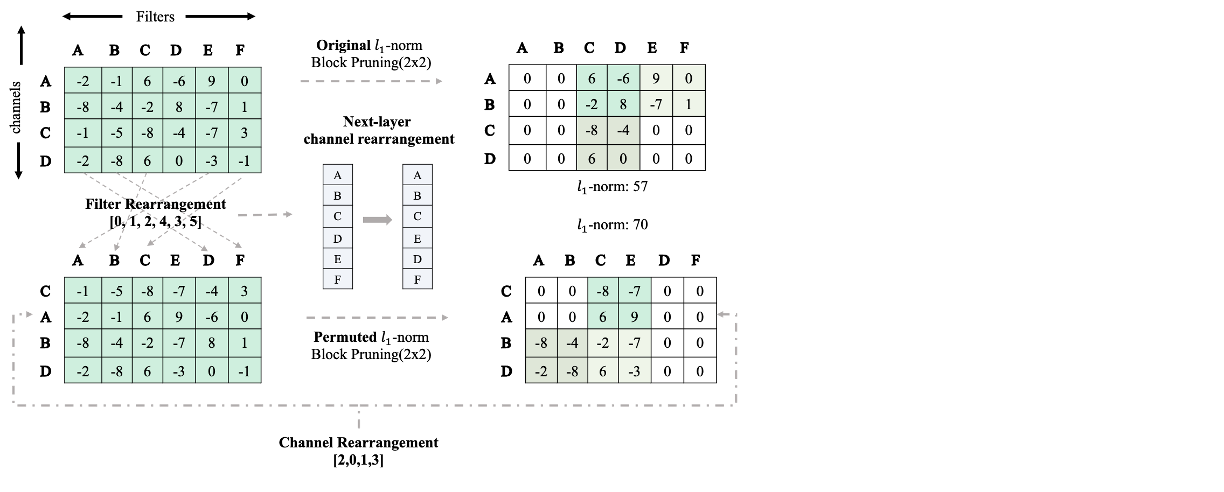}
	\caption{ Illustration of network rearrangement.  We rearrange the weight tensor by filter rearrangement index and channel rearrangement index to preserve weight magnitude.}
	\label{permute}
\vspace{-0.25em} 
\end{figure}

 \begin{figure*}[htbp]
	\centering
	\includegraphics[width=0.9\linewidth]{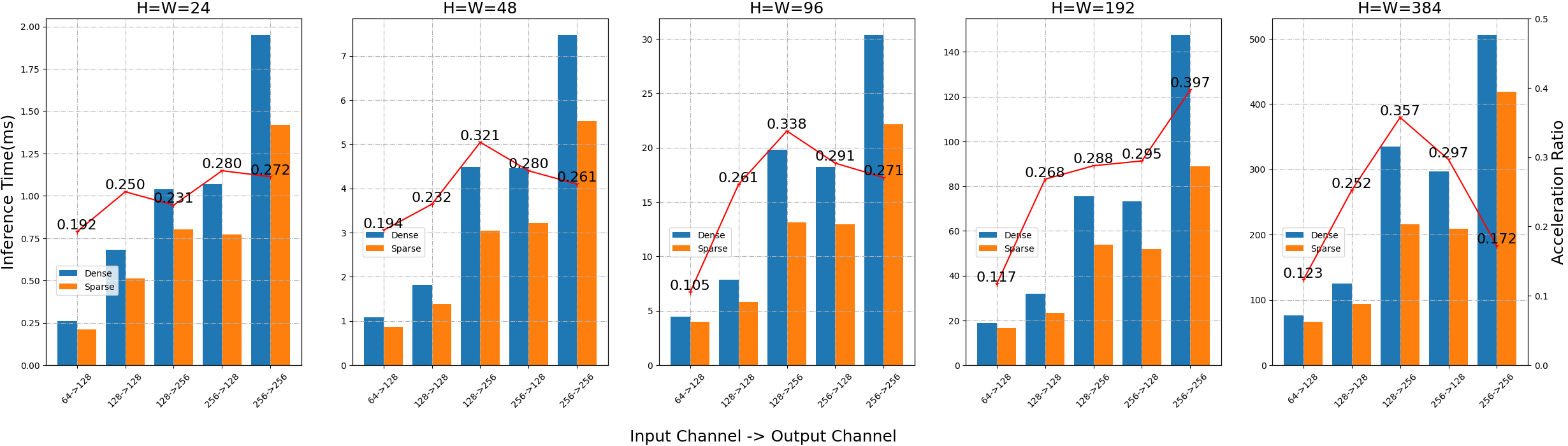}
	\caption{Acceleration performance of Conv1x1 under different configurations. The experiment is conducted on Qualcomm 855 CPU with 30\% sparse rate. Best view in colors.  }
	\label{conv1x1_cpu}
\end{figure*}

 \begin{figure*}[htbp]
	\centering
	\includegraphics[width=0.9\linewidth]{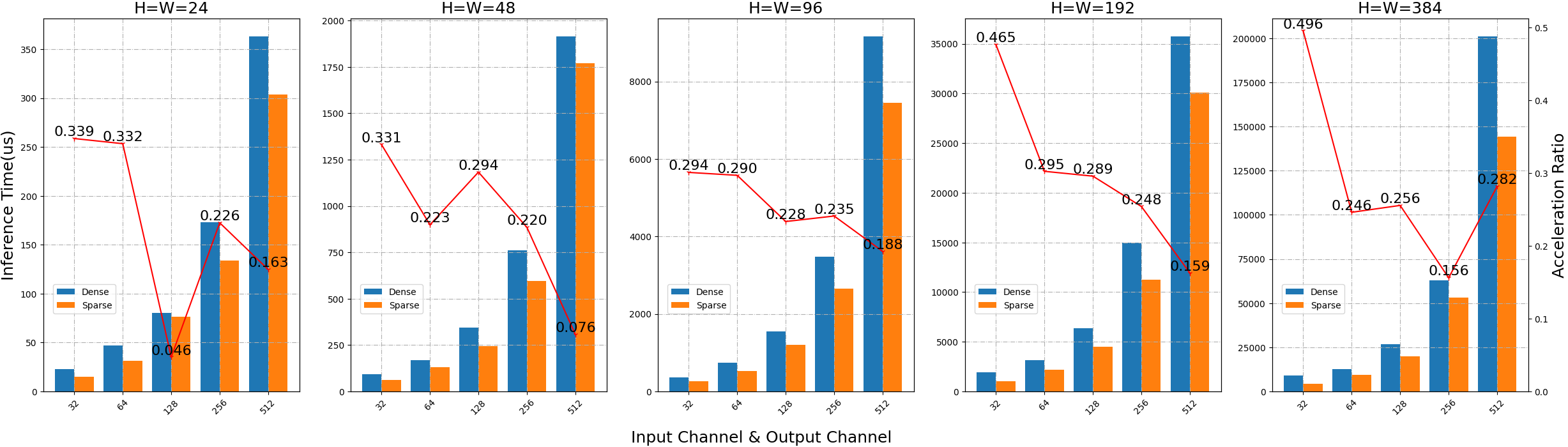}
	\caption{Acceleration performance of DwConv3x3 under different configurations. The experiment is conducted on Qualcomm 855 CPU with 33\% sparse rate. Best view in colors.}
	\label{dw_cpu}
\end{figure*}

\subsection{Whole network rearrangement}
\label{3.4}
For pattern group pruning and connectivity group pruning,  
we observed that when the importance of kernels in the group differs greatly, it will lead to evaluation inaccuracy, which means that relatively important kernels are affected by unimportant kernels. This observation motivates us to  change the layout of the weight tensor before pruning to reduce the importance variance of the kernels within a group. As shown in Fig \ref{permute}, we propose the whole network rearrangement strategy to derive more influential blocks for accuracy improvements. When the example matrix(top-left) is pruned by 50\% with a group size of $2\times2$,  it results in a sparse weight (top-right) with $l_{1}$-norm of 57. If we change the order of the input channel dimension and output channel dimension(bottom-left),  the resulting sparse weight (bottom-right) would have a total weight magnitude of 70. In order to avoid changing the output of the network, the rearrangement index needs to be propagated throughout the network graph, which means that the filter rearrangement index calculated by the "parent" layer will be used as the channel rearrangement index of "children" layers. 
Searching for good filter permutations for the target layer is challenging because for a layer with  $n^{i}$ filters, there exists $n^{i}!$ permutations, which is almost uncomputable for large $n^{i}$. 
However, the number of unique permutations can be reduced to $\frac{n^{i}!}{g_{o}!*(n^{i}/g_{o})!}$ in group pruning, in that both the order of filters in a group and the order of groups in a large matrix make no difference in accuracy improvements.  Each unique permutation can represent  $g_{o}!*(n^{i}/g_{o})!$ permutations, which will lead to the same sparse matrix $l1$-norm. To quickly search and evaluate unique permutations, we define a canonical form that
a permutation is unique only if each of its groups’ filters is in sorted order and the groups are sorted with respect to each other (e.g. by the first index value of each group). Then we use the bounded regression\cite{pool2021channel} method to quickly solve the above problem.

\subsection{Overview of  SparseByteNN Framework}
\label{3.5}
As shown in Fig\ref{framwork}, The classic neural network pruning process consists of three steps: training from scratch, pruning, and fine-traing. Before the second pruning step, we first rearrange the entire network to further reduce the pruning impact. Then, we apply FKGP pruning to obtain the sparse model and use fine-traing to recover the accuracy of the sparse model. Similar to NNI\cite{nni2021}, we encapsulate the above process into an algorithm compression component to provide users with out-of-the-box sparse fine-traing capabilities. The model conversion tool converts the ONNX model exported by the sparse fine-traing process into an sparse model IR\cite{vuduc2005fast}. Finally, based on the sparse model IR, the sparse inference engine completes the forward process  on the target hardware platform. 

\section{Experiments}
\label{4}
In this chapter, we first show that SparseByteNN has a better precision-speed trade-off than Filter Pruning, Weight Pruning, and other sparse engines in the industry through comparisons of different dimensions. Then, we prove the acceleration benefit of DwConv3x3 and Conv1x1, and the accuracy gain brought by the whole network rearrangement through a series of ablation. Finally, we extended FKGP to WebAssemebly and achieved remarkable performance.

\subsection{Implementation Settings}
\label{4.1}
In order to make the comparison fair and sufficient, we use the Filter Pruning and Weight Pruning algorithms contained in NNI\cite{nni2021} to construct a comparable experiment. The sparse rate is the real sparse rate of the entire network, which considers the interlayer coupling. All the experiments based on resnet20\cite{Idelbayev18a} of the CIFAR10\cite{krizhevsky2009learning} have the same hyperparameters, in which epochs, batch size, learning rate, and weight decay are set to 250, 128, 1e-2, and 1e-5 respectively, and the optimizer and scheduler are set to sgd\cite{bottou2012stochastic} and mstep respectively. Other experiments of ImageNet\cite{deng2009imagenet} are based on TIMM\cite{timm}. The pre-training and sparse-training of MobileNet-v1 use the same hyperparameters, in which epochs, batch size, learning rate, and weight decay are 300, 128, 0.045 and 1e-5 respectively, and the optimizer and scheduler are respectively choosing rmsproptf\cite{graves2013generating} and stepdecay, where decay-epochs is set to 2.4 and decay-rate is set to 0.973. 

\subsection{Performance Comparison}
\label{4.2}
\begin{figure}[htbp]  
	\centering
	\includegraphics[width=0.9\linewidth]{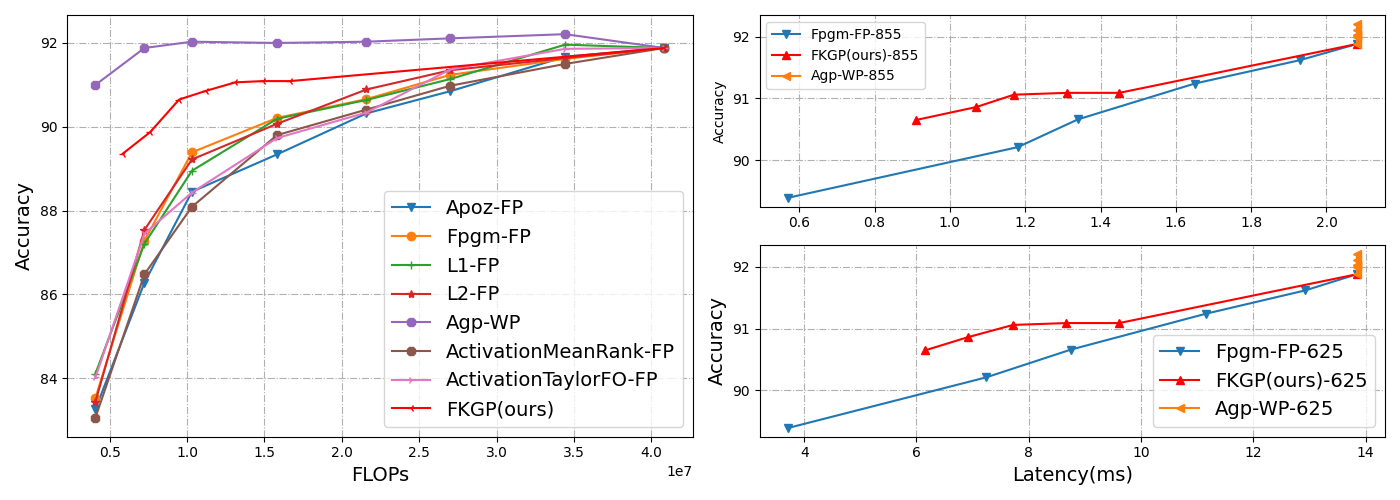}
	\caption{Performance comparison on resnet20(CIFAR10). 855 and 625 are Qualcomm chip models respectively }
	\label{resnet20_total}
\end{figure}
 
Firstly, we compare the performance of our FKGP with weight pruning and filter pruning on resnet20, which is stacked by Conv3x3 operators. As shown in Fig \ref{resnet20_total},  this comparison covers the  State-of-the-Art filter pruning, including Apoz\cite{apoz}, Fpgm\cite{fpgm}, L1\cite{l2}, L2\cite{l2},  ActivationMeanRank\cite{tylor}, ActivationTaylor\cite{tylor}, and weight pruning of Agp\cite{agp}.  Fig \ref{resnet20_total} shows that filter pruning suffers the most performance degradation and  the FKGP strategy surpasses all filter pruning. Although the accuracy of weight pruning under the same Flops exceeds that of FKGP, the former cannot obtain actual acceleration benefits. We conducted experiments on the actual latency of three types of pruning algorithms on mobile CPUs and found that FKGP exhibited a better speed-accuracy trade-off performance. Specifically, when the classification accuracy is 90.6\%, FKGP achieved a 34\% (0.91ms vs 1.34ms ) acceleration compared to FPGM on Qualcomm 855 and 29.6\%(6.16ms vs 8.75ms) on Qualcomm 625.

 \begin{table}[!htbp]
 \centering
  \caption{Performance comparison on MobileNet-v1(ImageNet) and the accuracy of the baseline model is 72.634\%}
 \label{tab::mbv1_compare}  
 \begin{tabular}
 {C{0.08\textwidth}C{0.09\textwidth}C{0.08\textwidth}C{0.08\textwidth}}
 	\toprule  
 	Methods& FLOPs(M) & Top-1 Acc(\%)$\uparrow$ & Latency (ms)\\
 	\midrule  
 	baseline &568& 0 &32.22\\
 	\midrule  
 	WP\cite{agp} &339& +0.022 &32.22\\
 	\midrule  
 	\multirow{5}*{FP\cite{fpgm}}  &511&-0.523&29.34\\
 	~&449& -1.102 & 25.34\\
 	~&397& -2.043& 22.41\\
 	~&339&-3.021& 19.33\\
 	~&284&-5.830& 16.04\\
 	\midrule  
 	\multirow{5}*{FKGP} &514&+0.568&30.56\\
 	    &460&+0.264& 27.96\\
 	    &406&-0.224& 25.29\\
 	    &352&-0.782& 22.68\\
 	    &299&-2.386& 19.68\\
 	\bottomrule 
 \end{tabular}
 \vspace{-0.5em}
\end{table}

Then, we compare the accuracy and latency on a lightweight neural network consisting of only conv1x1 and dwconv3x3. Since there is no obvious performance difference between different structured pruning algorithms, so we choose FPGM\cite{fpgm} as a representative. As shown in Table \ref{tab::mbv1_compare}, compared with the baseline MobileNet-v1\cite{mbv1}, when the pruning rate is 20\%, FKGP speeds up by 13\%, and the accuracy increases by 0.264\%. When the pruning rate is 40\%, it speeds up by 29.6\% while the accuracy only decreases by 0.78\%. Compared with the Filter pruning, the accuracy of FKGP has an advantage of 0.878\% when the latency is close to 25.3ms.

Finally, to further illustrate the acceleration advantages of SparseByteNN, taking mobilenetV1 as the baseline network and Qualcomm 855 as the test platform, we compare the performance between SparseByteNN and the SOTA on-mobile inference framework MNN\cite{Mnn}. Since MNN only supports sparse Conv1x1, for the fairness of the comparison, SparseByteNN turns off the sparse acceleration of DwConv3x3. As shown in Table \ref{tab::mnn_compare}, SparseByteNN is 3.21\% faster than MNN for the dense model. With the increase of the sparse rate, the performance advantage of SparseByteNN is further highlighted. When the sparse rate is 30\%, the performance advantage reaches a maximum of 22.30\%. Based on the identical experimental configuration, we obtained the accuracy at this sparse rate. The results show that although SparseByteNN has a larger sparse granularity, the accuracy drop is close to that of MNN(0.224\% vs 0.213\%). The experimental results and its technical documentation show that MNN will have a significant acceleration compared to the dense model only when the sparse rate reaches more than 30\%, and it suffers difficulties when generalized to DNN layers other than Conv1x1 layers. 
 \begin{table}[!htbp]
 \centering
  \caption{Comparison of inference time under different sparsity rates. Sp represents the sparse rate.}
 \label{tab::mnn_compare}  
 \resizebox{\linewidth}{!}{
 \begin{tabular}{c c c c c c}
 	\toprule  
 	Framework & Sp 0(ms) & Sp 0.1(ms) & Sp 0.2(ms) & Sp 0.3(ms) & Sp 0.5(ms) \\
 	\midrule  
 	MNN &33.29 & 33.37 &33.48 &32.55 & 20.23 \\
 	\midrule  
 	SparseByteNN &32.22 & 30.56 & 27.96 &25.29 & 19.68\\
 	\midrule  
    Speedup&3.21\% & 8.42\% & 16.48\% & 22.30\% & 2.71\%\\
 	\bottomrule 
 \end{tabular}
 }
 \vspace{-0.5em}
\end{table}

\subsection{Ablation study}
\label{4.3}
\subsubsection{Effectiveness of Sparse Patterns}
\label{4.3.1}

One of our main contributions is to design different pattern-based group pruning strategies for Conv1x1 and DwConv3x3 respectively, taking into account both accuracy and speed. We will prove the effectiveness of the fine-grained sparse model based on experiments.

\textbf{Conv1x1:} As described in Section \ref{3.2} and Section \ref{3.3}, we only perform connectivity group pruning with a group size of 4x4 on Conv1x1. Table \ref{tab::mbv1_compare} and Table \ref{tab::mnn_compare} show that SparseByteNN has performance advantages over SOTA pruning algorithms and sparse inference engines when only considering Conv1x1 pruning. In order to further illustrate the acceleration performance of the Conv1x1 operator, we conducted a comprehensive benchmark on common input configurations. As shown in Fig \ref{conv1x1_cpu}, when the sparsity rate is 30\%, the speedup of a single operator ranges from 11.50\% to 39.70\%, with a median of 26.80\% and a average of 25.38\%. Test results at more sparsity rates can be found in the appendix material.

 \begin{table}[!htbp]
 \centering
  \caption{Top1 accuracy comparison between the 3:9 patterns and the 5:9 patterns.}
 \label{tab::pattern_compare}  
 \resizebox{\linewidth}{!}{
 \begin{tabular}{c c c c c c}
 	\toprule  
 	Model & Params(M) & Baseline(\%) & 3:9 Patterns(\%) & 5:9 Patterns(\%)\\
 	\midrule  
 	MobileNet-v1 & 4.2 & 72.634  &72.954 &72.434  \\
 	\midrule  
 	MobileNet-v2 & 3.5 & 72.944 & 72.892 &72.256  \\
 	\midrule  
 	MobileNet-v3 & 5.5 & 75.776 & 75.703 & 74.973  \\
 	\bottomrule 
 \end{tabular}
 }
 \vspace{-0.5em}
\end{table}


\textbf{DwConv3x3:} To balance accuracy and speed, we only perform pattern group pruning with a group size of 1x16 for DwConv3x3. As shown in Table \ref{tab::pattern_compare}, compared with the 5:9 sparse patterns proposed by PatDNN, the 3:9 sparse patterns we designed for DwConv3x3 have lower accuracy loss and achieve approximately lossless pruning. Specifically, on MobileNet v1 to v3, the accuracy of 3:9 sparse exceeds 5:9 sparse by 0.52\%, 0.636\% and 0.73\% respectively. Furthermore, we perform benchmarks under common configurations to demonstrate the acceleration performance of a single operator. It should be noted that 3:9 sparse means that all DwConv3x3 have a fixed 33\% sparse rate. As shown in Fig \ref{dw_cpu}, when the sparsity rate is 33\%, the median speedup of a single operator is 24.8\%, and the average is 25.5\%. Under the configuration of a large feature map and small channels, the speedup can reach up to 49.6\%.

\subsubsection{Effectiveness of Network Rearrangement}
\label{4.3.2}

 \begin{table}[!htbp]
 \centering
  \caption{Performance studies with and without network rearrangement. Sp represents the sparse rate of Conv1x1, while DwConv3x3 is sparse.}
 \label{tab::rearrangement}  
 \resizebox{\linewidth}{!}{
 \begin{tabular}{c c c c c c}
 	\toprule  
 	Mobilent-V1 & Sp 0.1(\%) & Sp 0.2(\%) & Sp 0.3(\%) \\
 	\midrule  
 	w/o Rearrangement &72.692 &72.236  &71.74 \\
 	\midrule  
 	Rearrangement &72.970 &72.371 &72.124 \\
 	\bottomrule 
 \end{tabular}
 }
 \vspace{-0.5em}
\end{table}

In order to derive more influential blocks in group pruning, we propose the whole network rearrangement strategy. Table \ref{tab::rearrangement} presents the experimental results, which are conducted to further explore the impact of rearrangement under different Conv1x1 sparse rates on MobileNet-v1(ImageNet). From Table \ref{tab::rearrangement}, we observe that the whole network rearrangement can effectively improve network accuracy.

\subsection{WebAssemebly Acceleration}
\label{4.4}

\begin{figure}[htbp]  
	\centering
	\includegraphics[width=0.9\linewidth]{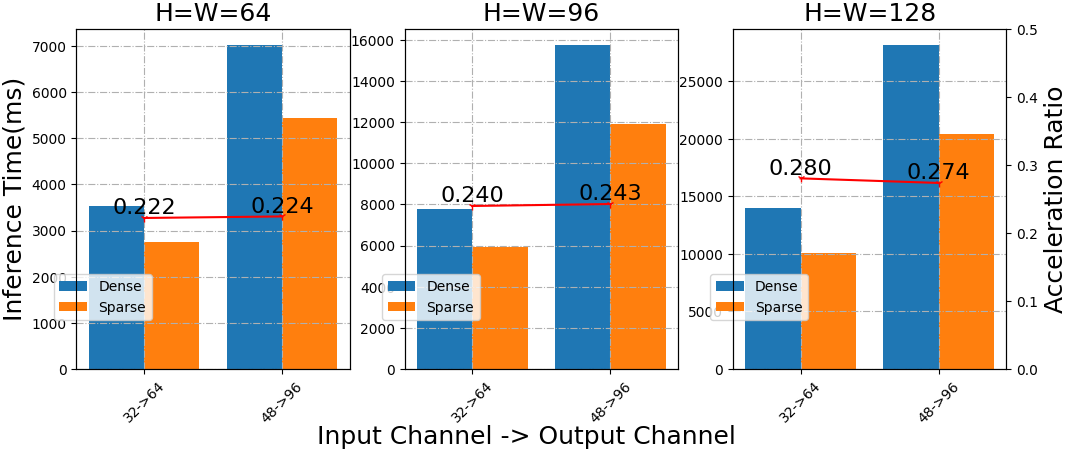}
	\caption{Speedup of Conv1x1 at 30\% sparse rate. The test platform is MacBook Pro 16, and the chrome version is 111.0.5563.64}
	\label{wasm}
\end{figure}

The above experimental results prove the excellent performance of our proposed fine-grained kernel group sparsity on ARM CPU. To illustrate the generalization of this strategy, we implemented efficient sparse kernels for Conv1x1 and DwConv3x3 based on WebAssembly, which can be used to accelerate neural network applications on the Web. As shown in Fig \ref{wasm}, when the input feature maps are 64x64, 96x96, and 128x128, under the channel configuration commonly used on the web side, the sparseness of 30\% can achieve an average speedup of 22.3\%, 24.15\%, and 27.7\%, respectively.Test results at more sparsity rates can be found in the appendix material.

\section{Conclusion and Future Work}
\label{4.5}
This work proposed a novel mobile inference acceleration framework named SparseByteNN, which provides end-to-end neural network acceleration capabilities from algorithms to engines on general CPUs. It contains a fine-grained group kernel sparsity schema and a family of co-optimized efficient sparse kernels. Combined with a customized network rearrangement strategy, SparseByteNN achieves real-time execution as well as high accuracy. The experiments on the MobileNets and CPUs platforms demonstrated that SparseByteNN has better speed and accuracy trade-off performance than the current SOTA pruning algorithms and sparse inference engine. In the future, we will further expand the application of pattern-based software-hardware collaborative sparse acceleration on more architectures, including mobile GPU(OpenCL) and server GPU(CUDA).
\section{Acknowledgement}
\label{4.6}

{\small
\bibliographystyle{ieee_fullname}
\bibliography{PaperForReview}

\begin{thebibliography}{10}\itemsep=-1pt

\bibitem{bottou2012stochastic}
L{\'e}on Bottou.
\newblock Stochastic gradient descent tricks.
\newblock {\em Neural Networks: Tricks of the Trade: Second Edition}, pages
  421--436, 2012.

\bibitem{TVM}
Tianqi Chen, Thierry Moreau, Ziheng Jiang, Lianmin Zheng, Eddie Yan, Meghan
  Cowan, Haichen Shen, Leyuan Wang, Yuwei Hu, Luis Ceze, et~al.
\newblock Tvm: An automated end-to-end optimizing compiler for deep learning.
\newblock {\em arXiv preprint arXiv:1802.04799}, 2018.

\bibitem{courbariaux2016binarized}
Matthieu Courbariaux, Itay Hubara, Daniel Soudry, Ran El-Yaniv, and Yoshua
  Bengio.
\newblock Binarized neural networks: Training deep neural networks with weights
  and activations constrained to+ 1 or-1.
\newblock {\em arXiv preprint arXiv:1602.02830}, 2016.

\bibitem{deng2009imagenet}
Jia Deng, Wei Dong, Richard Socher, Li-Jia Li, Kai Li, and Li Fei-Fei.
\newblock Imagenet: A large-scale hierarchical image database.
\newblock In {\em 2009 IEEE conference on computer vision and pattern
  recognition}, pages 248--255. Ieee, 2009.

\bibitem{deng2019deep}
Yunbin Deng.
\newblock Deep learning on mobile devices: a review.
\newblock In {\em Mobile Multimedia/Image Processing, Security, and
  Applications 2019}, volume 10993, pages 52--66. SPIE, 2019.

\bibitem{tflite}
Google.
\newblock Tensorflow lite.
\newblock \url{https://tensorflow.google.org/lite/}.

\bibitem{graves2013generating}
Alex Graves.
\newblock Generating sequences with recurrent neural networks.
\newblock {\em arXiv preprint arXiv:1308.0850}, 2013.

\bibitem{gupta2015deep}
Suyog Gupta, Ankur Agrawal, Kailash Gopalakrishnan, and Pritish Narayanan.
\newblock Deep learning with limited numerical precision.
\newblock In {\em International conference on machine learning}, pages
  1737--1746. PMLR, 2015.

\bibitem{ghostnet}
Kai Han, Yunhe Wang, Qi Tian, Jianyuan Guo, Chunjing Xu, and Chang Xu.
\newblock Ghostnet: More features from cheap operations.
\newblock In {\em Proceedings of the IEEE/CVF conference on computer vision and
  pattern recognition}, pages 1580--1589, 2020.

\bibitem{deep}
Song Han, Huizi Mao, and William~J Dally.
\newblock Deep compression: Compressing deep neural networks with pruning,
  trained quantization and huffman coding.
\newblock {\em arXiv preprint arXiv:1510.00149}, 2015.

\bibitem{learning}
Song Han, Jeff Pool, John Tran, and William Dally.
\newblock Learning both weights and connections for efficient neural network.
\newblock {\em Advances in neural information processing systems}, 28, 2015.

\bibitem{resnet}
Kaiming He, Xiangyu Zhang, Shaoqing Ren, and Jian Sun.
\newblock Deep residual learning for image recognition.
\newblock In {\em Proceedings of the IEEE conference on computer vision and
  pattern recognition}, pages 770--778, 2016.

\bibitem{he2018soft}
Yang He, Guoliang Kang, Xuanyi Dong, Yanwei Fu, and Yi Yang.
\newblock Soft filter pruning for accelerating deep convolutional neural
  networks.
\newblock {\em arXiv preprint arXiv:1808.06866}, 2018.

\bibitem{fpgm}
Yang He, Ping Liu, Ziwei Wang, Zhilan Hu, and Yi Yang.
\newblock Filter pruning via geometric median for deep convolutional neural
  networks acceleration.
\newblock In {\em Proceedings of the IEEE/CVF conference on computer vision and
  pattern recognition}, pages 4340--4349, 2019.

\bibitem{mbv1}
Andrew~G Howard, Menglong Zhu, Bo Chen, Dmitry Kalenichenko, Weijun Wang,
  Tobias Weyand, Marco Andreetto, and Hartwig Adam.
\newblock Mobilenets: Efficient convolutional neural networks for mobile vision
  applications.
\newblock {\em arXiv preprint arXiv:1704.04861}, 2017.

\bibitem{apoz}
Hengyuan Hu, Rui Peng, Yu-Wing Tai, and Chi-Keung Tang.
\newblock Network trimming: A data-driven neuron pruning approach towards
  efficient deep architectures.
\newblock {\em arXiv preprint arXiv:1607.03250}, 2016.

\bibitem{l2}
Zhongzhan Huang, Wenqi Shao, Xinjiang Wang, Liang Lin, and Ping Luo.
\newblock Rethinking the pruning criteria for convolutional neural network.
\newblock {\em Advances in Neural Information Processing Systems},
  34:16305--16318, 2021.

\bibitem{Idelbayev18a}
Yerlan Idelbayev.
\newblock Proper {ResNet} implementation for {CIFAR10/CIFAR100} in {PyTorch}.
\newblock \url{https://github.com/akamaster/pytorch_resnet_cifar10}.
\newblock Accessed: 20xx-xx-xx.

\bibitem{Mnn}
Xiaotang Jiang, Huan Wang, Yiliu Chen, Ziqi Wu, Lichuan Wang, Bin Zou, Yafeng
  Yang, Zongyang Cui, Yu Cai, Tianhang Yu, et~al.
\newblock Mnn: A universal and efficient inference engine.
\newblock {\em Proceedings of Machine Learning and Systems}, 2:1--13, 2020.

\bibitem{krizhevsky2009learning}
Alex Krizhevsky, Geoffrey Hinton, et~al.
\newblock Learning multiple layers of features from tiny images.
\newblock {\em cs.utoronto.ca}, 2009.

\bibitem{lane2016deepx}
Nicholas~D Lane, Sourav Bhattacharya, Petko Georgiev, Claudio Forlivesi, Lei
  Jiao, Lorena Qendro, and Fahim Kawsar.
\newblock Deepx: A software accelerator for low-power deep learning inference
  on mobile devices.
\newblock In {\em 2016 15th ACM/IEEE International Conference on Information
  Processing in Sensor Networks (IPSN)}, pages 1--12. IEEE, 2016.

\bibitem{lane2017squeezing}
Nicholas~D Lane, Sourav Bhattacharya, Akhil Mathur, Petko Georgiev, Claudio
  Forlivesi, and Fahim Kawsar.
\newblock Squeezing deep learning into mobile and embedded devices.
\newblock {\em IEEE Pervasive Computing}, 16(3):82--88, 2017.

\bibitem{lecun1989optimal}
Yann LeCun, John Denker, and Sara Solla.
\newblock Optimal brain damage.
\newblock {\em Advances in neural information processing systems}, 2, 1989.

\bibitem{l1}
Hao Li, Asim Kadav, Igor Durdanovic, Hanan Samet, and Hans~Peter Graf.
\newblock Pruning filters for efficient convnets.
\newblock {\em arXiv preprint arXiv:1608.08710}, 2016.

\bibitem{lin2020dynamic}
Tao Lin, Sebastian~U Stich, Luis Barba, Daniil Dmitriev, and Martin Jaggi.
\newblock Dynamic model pruning with feedback.
\newblock {\em arXiv preprint arXiv:2006.07253}, 2020.

\bibitem{slim}
Zhuang Liu, Jianguo Li, Zhiqiang Shen, Gao Huang, Shoumeng Yan, and Changshui
  Zhang.
\newblock Learning efficient convolutional networks through network slimming.
\newblock In {\em Proceedings of the IEEE international conference on computer
  vision}, pages 2736--2744, 2017.

\bibitem{luo2020comparison}
Chunjie Luo, Xiwen He, Jianfeng Zhan, Lei Wang, Wanling Gao, and Jiahui Dai.
\newblock Comparison and benchmarking of ai models and frameworks on mobile
  devices.
\newblock {\em arXiv preprint arXiv:2005.05085}, 2020.

\bibitem{pytorch-mobile}
Meta.
\newblock Pytorch mobile.
\newblock \url{https://pytorch.org/mobile/home/}.

\bibitem{nni2021}
{Microsoft}.
\newblock {Neural Network Intelligence}, 1 2021.

\bibitem{24}
Asit Mishra, Jorge~Albericio Latorre, Jeff Pool, Darko Stosic, Dusan Stosic,
  Ganesh Venkatesh, Chong Yu, and Paulius Micikevicius.
\newblock Accelerating sparse deep neural networks.
\newblock {\em arXiv preprint arXiv:2104.08378}, 2021.

\bibitem{tylor}
Pavlo Molchanov, Arun Mallya, Stephen Tyree, Iuri Frosio, and Jan Kautz.
\newblock Importance estimation for neural network pruning.
\newblock In {\em Proceedings of the IEEE/CVF Conference on Computer Vision and
  Pattern Recognition}, 2019.

\bibitem{admm}
Pavlo Molchanov, Arun Mallya, Stephen Tyree, Iuri Frosio, and Jan Kautz.
\newblock Importance estimation for neural network pruning.
\newblock In {\em Proceedings of the IEEE/CVF Conference on Computer Vision and
  Pattern Recognition}, pages 11264--11272, 2019.

\bibitem{patDNN}
Wei Niu, Xiaolong Ma, Sheng Lin, Shihao Wang, Xuehai Qian, Xue Lin, Yanzhi
  Wang, and Bin Ren.
\newblock Patdnn: Achieving real-time dnn execution on mobile devices with
  pattern-based weight pruning.
\newblock In {\em Proceedings of the Twenty-Fifth International Conference on
  Architectural Support for Programming Languages and Operating Systems}, pages
  907--922, 2020.

\bibitem{pool2021channel}
Jeff Pool and Chong Yu.
\newblock Channel permutations for n: m sparsity.
\newblock {\em Advances in Neural Information Processing Systems},
  34:13316--13327, 2021.

\bibitem{mbv2}
Mark Sandler, Andrew Howard, Menglong Zhu, Andrey Zhmoginov, and Liang-Chieh
  Chen.
\newblock Mobilenetv2: Inverted residuals and linear bottlenecks.
\newblock In {\em Proceedings of the IEEE conference on computer vision and
  pattern recognition}, pages 4510--4520, 2018.

\bibitem{vgg}
Karen Simonyan and Andrew Zisserman.
\newblock Very deep convolutional networks for large-scale image recognition.
\newblock {\em arXiv preprint arXiv:1409.1556}, 2014.

\bibitem{inceptionv4}
Christian Szegedy, Sergey Ioffe, Vincent Vanhoucke, and Alexander~A Alemi.
\newblock Inception-v4, inception-resnet and the impact of residual connections
  on learning.
\newblock In {\em Thirty-first AAAI conference on artificial intelligence},
  2017.

\bibitem{efficientnet}
Mingxing Tan and Quoc Le.
\newblock Efficientnet: Rethinking model scaling for convolutional neural
  networks.
\newblock In {\em International conference on machine learning}, pages
  6105--6114. PMLR, 2019.

\bibitem{vuduc2005fast}
Richard~W Vuduc and Hyun-Jin Moon.
\newblock Fast sparse matrix-vector multiplication by exploiting variable block
  structure.
\newblock In {\em High Performance Computing and Communications: First
  International Conference, HPCC 2005, Sorrento, Italy, September 21-23, 2005.
  Proceedings 1}, pages 807--816. Springer, 2005.

\bibitem{timm}
Ross Wightman.
\newblock Pytorch image models.
\newblock \url{https://github.com/rwightman/pytorch-image-models}, 2019.

\bibitem{auto-patdnn}
Zheng Zhan, Yifan Gong, Pu Zhao, Geng Yuan, Wei Niu, Yushu Wu, Tianyun Zhang,
  Malith Jayaweera, David Kaeli, Bin Ren, et~al.
\newblock Achieving on-mobile real-time super-resolution with neural
  architecture and pruning search.
\newblock In {\em Proceedings of the IEEE/CVF International Conference on
  Computer Vision}, pages 4821--4831, 2021.

\bibitem{agp}
Michael Zhu and Suyog Gupta.
\newblock To prune, or not to prune: exploring the efficacy of pruning for
  model compression.
\newblock {\em arXiv preprint arXiv:1710.01878}, 2017.

\end{thebibliography}
}

\end{document}